# Large language models are not about natural language

Johan J. Bolhuis[a]*, Andrea Moro[b,c], Stephen Crain[d] and Sandiway Fong[e]

[a]University of Cambridge, Department of Psychology, Downing Street, Cambridge CB2 3EB, UK.
[b]University School for Advanced Studies, Pavia, Italy
[c]Scuola Normale Superiore, Pisa, Italy.
[d]Macquarie University, Department of Linguistics, Sydney, Australia.
[e]University of Arizona, Department of Linguistics, Tucson, Arizona, USA.

*corresponding author. Email: jjb19@cam.ac.uk

## Abstract

Large Language Models are useless for linguistics, as they are probabilistic models that require a vast amount of data to analyse externalized strings of words. In contrast, human language is underpinned by a mind-internal computational system that recursively generates hierarchical thought structures. The language system grows with minimal external input and can readily distinguish between real language and impossible languages.

The title of the target article (Futrell & Mahowald, 2025) is a little confusing, particularly with regard to what is meant by 'language models'. In any science, the starting point for research is the formulation of a model, a theory of the underlying mechanisms that could go towards explaining the particular phenomenon at hand. In linguistics, the Strong Minimalist Thesis (Chomsky, 2000) is such a theory, a model for the mechanism of language. It becomes clear from the target article itself, that Futrell and Mahowald specifically refer to Large Language Models (LLMs). We disagree with all the arguments that the authors offer to support their claim that LLMs can be useful for our understanding of "linguistic structure, language processing, and learning." This is because human language is a computational system that works in a fundamentally different way from LLMs (Moro et al., 2023; Bolhuis et al., 2023, 2024; Fong, 2025). LLMs are probabilistic models that analyse externalized—'flattened'—strings of words. As such, they are not new: already in 1913, Markov (1913/2006) published a probabilistic analysis of vowels and consonants confirming a atrong bias in favor of an alternating vowel/consonant pattern in the poetry of Pushkin's *Eugene Onegin*. The probabilistic nature of LLMs is in complete contrast to the recursive function by which the human mind generatively forms hierarchical thought structures that may or may not be externalizable (Everaert et al., 2015; Friederici et al., 2017; Fong, 2025). The nature of those structures determines meaning (semantics) (Everaert et al., 2015). LLMs are not generative in this sense, and they do not form structures that determine meaning.

In addition, LLMs 'acquire' language in a fundamentally different way to human infants, who can build syntactic structures in their mind even without any relevant input (Crain et al., 2017; Yang et al., 2017; Bolhuis et al., 2023)—the 'Poverty of the Stimulus' argument that the target authors discuss. This is in stark contrast to the enormous amount of training plus trillion-parameter models that is absolutely required for LLMs to generate natural output (Fong, 2025). The contrast between the two is aptly illustrated by the differences in energy use. Although is it not often made public, we know that the environmental impact and power requirements of the latest language models are quite staggering. For example, it has been reported that xAI in Memphis TN requires 70MW to run 100,000 GPUs concurrently, power that the local utility was initially unable to supply. As a result, 18 natural gas generators (2.5MW each) had to be parked outside (Kerr, 2024). Similarly, Google has ordered six or seven small nuclear reactors for its AI datacenters according to Lawson (2024). In contrast, it is estimated that the human brain consumes about 20W, not all of which is available to language (Ling, 2001).

The manner in which LLMs learn linguistic patterns is also quite unlike the developmental stages children go through in the transition to adult language. At some stages, children's production and understanding of language systematically differs from that of adults, and includes linguistic structures that are not attested in the adult input (Crain, 2012). The stages of child language development have no semblance to the way LLMs assimilate linguistic patterns from training regimes. LLMs therefore shed no light on the cognitive architecture that enables human children to acquire languages.

Unlike human brains, machines based on LLMs are unable to distinguish between possible and impossible languages (Moro, 2016; Moro et al., 2023; Bolhuis et al., 2024). That the human brain can make this distinction is based on convergent experiments, exploiting both actual languages (Musso et al., 2003) and invented languages (Tettamanti et al., 2002). The common paradigm consists in contrasting rules based on linear sequences of words vs. hierarchical constituents, which are the ones based in all and only human languages (Rizzi, 2009) showing a selective inhibition of a network involving Broca's area for impossible languages only. The target authors address this issue, and argue that LLMs have "inductive biases [that] meaningfully align with linguistic structure". They specifically refer to a study of their own (Kallini et al., 2024), that purports to show that LLMs learn from English text significantly better than from impossible variants. However, the authors fail to mention other work that has shown that their conclusions are unwarranted. As Bowers (2025) pointed out, the worst performance of the LLMs in Kallini et al. (2024) was with a 'language' where the words were shuffled randomly within sentences. As Bowers (2025) rightly concludes, this has no structure at all, and thus is not a language, impossible or otherwise. The other examples with structured languages resulted in minimal or no differences between LLMs and human learners (see also Mitchell & Bowers, 2020). What differences there were could be explained by (non-linguistic) computational complexity of the 'impossible' languages used. As Bowers (2025) puts it, "What makes a language difficult to learn is not the same for LLMs and humans, because they possess different inductive biases.". More recently, Luo et al. (2024) and Ziv et al. (2025) found that LLMs did not distinguish between normal English and backward reversed English text, again suggesting that, unlike the human language faculty (Moro, 2016) they make no distinction between possible and impossible languages.

The observation that LLMs learn to simulate impossible languages just as well as possible languages renders LLMs uninformative about the cognitive structures children use to acquire language. For one thing, children are incapable of acquiring linguistic patterns based on linear order, as Noam Chomsky has pointed out for decades (e.g., Chomsky, 1971). By contrast, the whole point about LLMs is that they can readily detect linear regularities in language input. This shows that LLMs and human children process language input in fundamentally different ways.

Given these fundamental differences between LLMs and the human language faculty, we fail to see how the former can inform the latter in any meaningful way. In summary, an LLM may quack like a duck, but isn't one—and, in its current probabilistic form, it can never be one.